\documentclass{article}
\usepackage{spconf,amsmath,graphicx}

\title{Using previous acoustic context to improve Text-to-Speech synthesis}

\name{Pilar Oplustil Gallegos, Simon King\thanks{This work was supported by ANID, Becas Chile scholarship nº 72190135.}}
\address{The Centre for Speech Technology Research\\ University of Edinburgh\\ United Kingdom}
\begin{document}
%
\maketitle
\begin{abstract} 
Many speech synthesis datasets, especially those derived from audiobooks, naturally comprise sequences of utterances. Nevertheless, such data are commonly treated as individual, unordered utterances both when training a model and at inference time. This discards important prosodic phenomena above the utterance level. In this paper, we leverage the sequential nature of the data using an acoustic context encoder that produces an embedding of the previous utterance audio. This is input to the decoder in a Tacotron 2 model. The embedding is also used for a secondary task, providing additional supervision. We compare two secondary tasks: predicting the ordering of utterance pairs, and predicting the embedding of the current utterance audio. Results show that the relation between consecutive utterances is informative: our proposed model significantly improves naturalness over a Tacotron 2 baseline.
\end{abstract}
\section{Introduction}

Some of the common sources of data to train state-of-the-art machine learning Text-to-Speech (TTS) models include audiobooks and other extended monologues. The raw data, comprised of large coherent speech units, such as paragraphs, is pre-processed into utterances (approximately sentences). At training time, a model is presented with the resulting paired speech-text utterances, treated as independent from each other, discarding the original ordering. With current neural architectures, hardware memory limitations necessitate this pre-processing into utterances. But, it disregards important information available in the original speech data, such as long-form prosodic patterns. 

Prosodic variation is governed by context at different levels, and contextual information is expressed through prosody \cite{cole2015prosody}. For example, in monologues, utterances exhibit significant prosodic differences depending on their position in longer units (e.g. transcribed paragraphs) \cite{farrus2016paragraph}. Patterns include: decrease of fundamental frequency through time; speech rate peaks around the middle of the larger unit. In dialogue, prosody is affected by the interaction. Specific prosodic cues have been observed in utterances where speakers change turn, such as falling and high-rising final intonations and increased pause duration \cite{gravano2011turn}. 

TTS systems might be deployed in dialogue agents or audiobook readers where, to obtain natural and appropriate prosody, it would be desirable to reproduce these prosodic phenomena. However, pre-processing that discards all knowledge of the structure of the original speech data seems likely to make it harder to accurately predict such prosodic behaviour. 

In this paper, we make a first step towards modelling longer speech phenomena by proposing a model that uses acoustic context. We think about a speech dataset as a single acoustic sequence where, inside a long speech unit such as an audiobook chapter, utterance $N$ depends on the previous utterance $N-1$ (and, potentially, as many previous utterances as available). Our method incorporates this ``$N$-gram'' idea into the training procedure of a sequence-to-sequence TTS model. 

While the \textit{evaluation} of TTS has already been criticized for working mostly with isolated sentences \cite{clark2019evaluating}, little has been done in this respect about training. It can be argued that, for monologues, the fact that we can't use longer portions of speech is limited \textit{only} by memory and computing restrictions. However, a TTS system for interactive dialogue should be able to model the dependency between user utterances and the system ones, in a real-time, causal fashion. This implies that only previous (i.e., past) context should be used.

Our contributions are: (1) we develop a method to use previous acoustic context in a sequence-to-sequence (S2S) TTS system; (2) we show, through subjective evaluation, that this method improves the general synthetic speech naturalness; (3) we show initial results for the perceived preference of our system output when heard in-context.

\vfill
\section{Related work}

Work in earlier TTS frameworks tried to incorporate prosody above the utterance level by including additional categorical information derived from the text. For HMM-TTS synthesis, \cite{peiro2018paragraph} incorporates inter- and intra-paragraph positional features that are used to modify prosody through rules to fit to the patterns analyzed in \cite{farrus2016paragraph}. In \cite{klimkov2017phrase} several text-based features to predict phrase breaks for long-form reading TTS were tested. Word embeddings yield the best results, when used to train a model to predict prosodic phrase breaks occurring without punctuation.

More recently the focus has shifted to using acoustic information either to predict categorical information or to directly enhance the models. In \cite{suni2020prosodic}, prominence and boundary labels are generated from the speech signal using a continuous wavelet transform method, and this information is then used to train a S2S TTS system. They showed through objective evaluation that, while the baseline system can place prominence labels at correct positions in the utterance, it fails to accurately predict the strengths; their model improves on this. However, as the authors admit, at synthesis time these labels would need to be predicted from text.

In \cite{wang2018style} Global Style Tokens (GSTs) were proposed to enhance the state-of-the-art model Tacotron. The sequence of Mel spectrum frames for a training sample are input into a reference encoder which extracts a reference embedding. An attention module learns a similarity measure between the reference embedding and a number of tokens shared across all training data. The weights for each token are combined to form a `style' embedding. At synthesis time, the model can either be provided with those weights or derive them from a reference speech signal, whose global `style' characteristics will be transferred to the synthesized speech. The model doesn't minimise any additional loss and doesn't require any external labels. 

In \cite{tyagi2019dynamic}, the authors develop a method to sample embeddings from a Variational Auto-Encoder (VAE) for multi-sentence speech synthesis, by taking into account the acoustic distance to the previous utterance, among other factors. They transform the utterance embedding space into a 2D space through Principal Component Analysis and measure the acoustic similarity by using Euclidean distance. This idea is somewhat similar to our proposed system, but they only apply it at synthesis time, while we propose to use acoustic context during training in order to exploit knowledge of the structure of the original speech data.

While the idea of using extra acoustic information at training and synthesis time is not new in the TTS field, to our knowledge it hasn't been used to encode contextual information at training time. 

We do find examples of this concept in other Deep Learning fields. For attention-based neural machine translation, \cite{tiedemann2017neural} proposes to train with an extended context. They show that its use improves the translation of phenomena that require discourse level information, such as pronoun disambiguation. Their proposed system incorporates the use of source language sentences of previous translation units, which are automatically labelled with a special context token. They analyze the behaviour of the attention over the context words, and they see that they are attended about 7.1\% of the time. 

For natural language understanding, \cite{jernite2017discourse} trains sentence encoders that incorporate paragraph-level discourse information through a multi-task framework. They propose to train a double sentence encoder with tied parameters to process pairs of sentences. Additionally, they design three discourse-inspired objectives, two of them directly relevant for us: (1) a \textit{order classifier}, where the order of the paired sentences is switched with a probability of 0.5, and the model must classify if the original order has been kept or not; (2) a \textit{next classifier}, where the model is presented with a set of candidates and needs to classify which one corresponds to the next sentence in the pair. The main advantage of their proposed tasks is that they don't require labelled data: it is the ordering of sentences itself that is used as additional supervision for learning. The idea in \cite{tiedemann2017neural} could be directly applicable to TTS, but memory considerations need to be met. Therefore in this paper we propose an approach more similar to \cite{jernite2017discourse}: we introduce additional tasks into the TTS model which require the model to account for acoustic context.

\section{System Description}


\subsection{Leveraging consecutive utterances}

We propose to model the inherently sequential nature of speech utterances in monologue. TTS datasets are usually pre-processed into sentences, utterances, or other more- or less-well-defined chunks that, depending on how the data was elicited, belong to larger speech units. We will use the term \textit{utterance} to refer to these chunks. For example, in audiobooks, utterances occur in a linear order; that information is usually available. 

We propose to modify the training scheme such that generation of the current utterance ($N$) is conditioned on the acoustics of the previous utterance ($N-1$), given the order in the dataset: for example, two consecutive utterances in an audiobook. This information cannot in general be easily inferred from the text, and is therefore additional information for the model. We will require the model to learn the relationship between consecutive utterances, i.e., utterance bigrams. 

\subsection{Acoustic Context Encoder}

To incorporate the additional acoustic context (i.e., utterance $N-1$) into the model, we add an Acoustic Context Encoder (ACE) to a Tacotron 2 \cite{shen2018natural} baseline architecture. There are many available architectural options to encode the utterance. In this work, we made use of the already available GST architecture. As shown in the original paper, this module can encode the prosodic characteristics of any given utterance, and provide a fixed sized embedding that can be concatenated to Tacotron's text encoder outputs. In the original paper, the input to the GST module during training is the mel-scale spectrogram of the current utterance ($N$). In contrast, in our proposed system, we separately encode utterances $N$ and $N-1$ and use these two encodings for an additional task described in the next section.

\subsection{Multi-task training}

\begin{figure}
    \centering
    \includegraphics[width=0.45\textwidth]{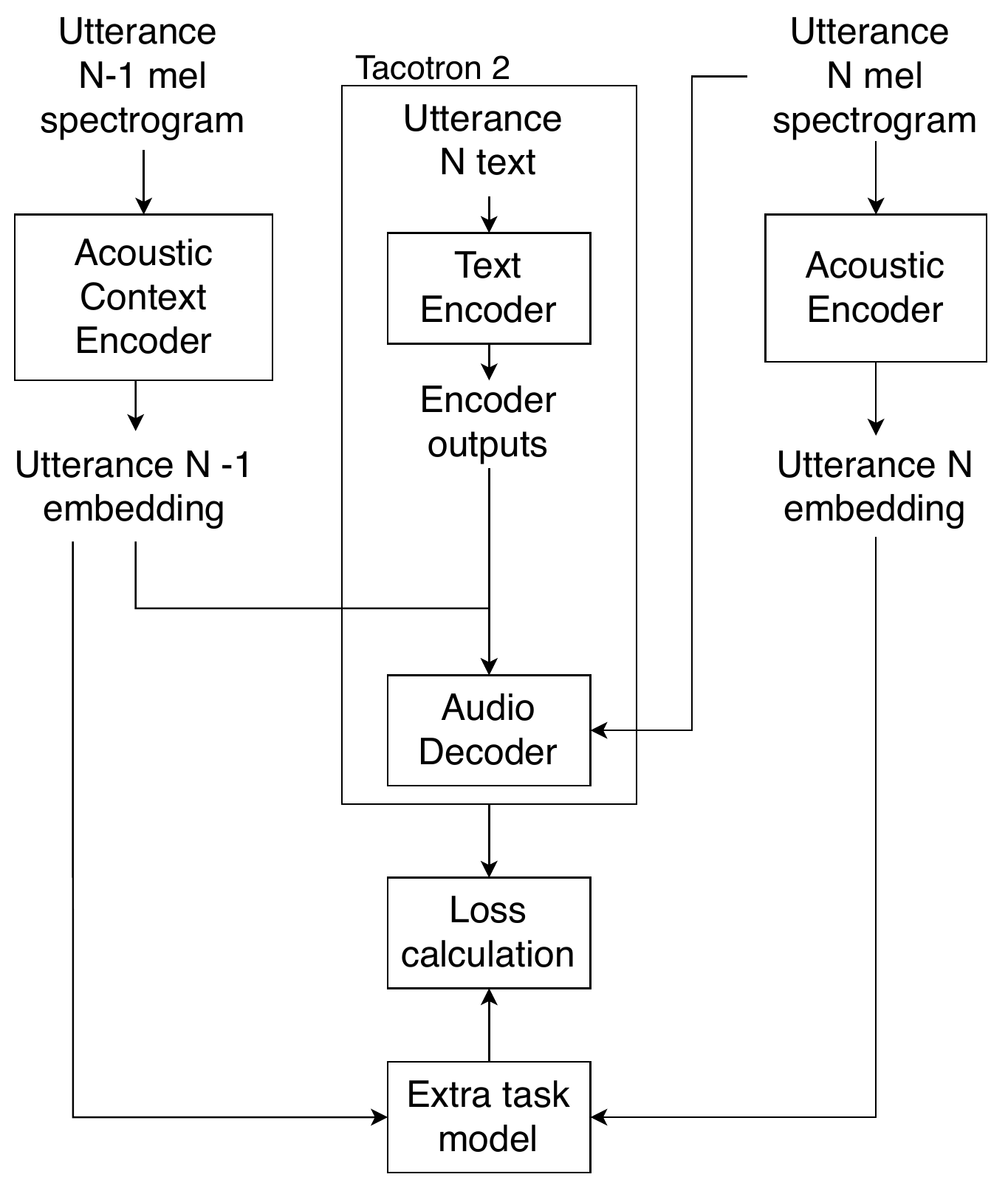}
    \caption{Diagram of the proposed system. For example, using the LJ speech dataset, when utterance $N-1$ corresponds to sample LJ026-0047, utterance $N$ corresponds to LJ026-0048.}
    \label{fig:system}
\end{figure}

To require the model to learn the bigram relationship between consecutive utterances, we propose a multi-task training framework. We use encodings of utterance $N-1$ and of utterance $N-1$ as input to an additional task model, and minimise the loss on that task in addition to the usual Tacotron 2 loss for the main TTS task. Both the Acoustic Context Encoder, fed with utterance $N-1$, and the Acoustic Encoder, fed with utterance $N$, are instances of the GST architecture, however they don't share parameters. See Figure \ref{fig:system} for the complete proposed system. We experiment with two tasks:

\begin{enumerate}
    \item \textbf{Order Task} As in \cite{jernite2017discourse}, the extra task is binary classification: predict the order of the two utterance embeddings. During training, randomly with a probability of 0.5, these are presented in original or in reversed order. The classifier must predict if the order is correct. A binary loss is added to the main Tacotron 2 loss.
    \item \textbf{Next Task} the extra task is regression: predict the embedding of utterance $N$ given the embedding of utterance $N-1$. An MSE loss is added to the main Tacotron 2 loss. Although this not the same task as  presented in \cite{jernite2017discourse}, we use the same name because it was inspired by their design.
\end{enumerate}

In both cases, the encoding of the current utterance ($N$) is only used at training time because it is needed to calculate the extra task loss. Unlike in the original GST model, it is not a reference encoder; it is not needed at synthesis time, when only the Acoustic Context Encoder (ACE) is required.

We also created a model denoted \textbf{ACE only} which comprises the Tacotron 2 model plus the Acoustic Context Encoder for utterance $N-1$, with its output embedding being fed to the input of the Tacotron 2 decoder as in Figure \ref{fig:system}. This system has access to acoustic context, but only as an additional input. It does not perform and extra task, and is trained only to minimise the usual Tacotron 2 loss for the TTS task.
 
\subsection{Data and Tools}

We use the LJ speech dataset \cite{ljspeech17},  comprised of seven books, pre-processed into utterances by automatically chunking at silences. Therefore, during training, the utterance \textit{chapter 01 chunk 03} is conditioned on the utterance \textit{chapter 01 chunk 02}, and this ordering information is easily available in the dataset metadata. We use a split of 12604 / 311 / 290 utterances for train / validation / test, for each subset.

Our \textbf{Baseline} TTS system is Tacotron 2 \cite{shen2018natural}. We use the NVIDIA \footnote{https://github.com/NVIDIA/tacotron2} code to which we added the GST implementation from their Mellotron \cite{valle2020mellotron} code\footnote{https://github.com/NVIDIA/mellotron}. Hyperparameters were set to the default values given in the repositories and not adjusted. 

All models were trained from scratch for 100k iterations, a point where the validation loss had converged and the mel spectrogram quality through the neural vocoder was stable. We used a pretrained Waveglow neural vocoder \cite{prenger2019waveglow} \footnote{https://github.com/NVIDIA/waveglow checkpoint ``waveglow 256 channels universal v4''}.

The \textbf{Order Task} binary classification model consisted of 4 feed-forward layers of the following input and output dimensions: [512, 256], [256, 128], [128, 64] and [64, 1] with ReLU activation functions, batch normalization and dropout before the output layer. For the \textbf{Next Task} the regression model consisted of 5 layers of the following input and output dimensions: [256, 128], [128, 64], [64, 64], [64, 128] and [128, 256], with ReLU activation functions, batch normalization and dropout after the middle layer.

\section{Evaluation and Results}

Although our motivation was to better model  prosody above the utterance level, our proposed system should improve the general quality of the synthesized speech. In fact, we expect it to improve the perceived quality of utterance $N$ heard not only in context (i.e., after hearing utterance $N-1$), but also when heard in \textit{isolation}. This is because, during training, information from the acoustics of utterance $N-1$ will help the model explain some of the variation in utterance $N$ that would otherwise be unexplained if given only the text of utterance $N$. Therefore our evaluation is aimed at general TTS quality and not specifically to evaluate prosody.

We divided our evaluation of naturalness into two stages\footnote{Find all the samples in: https://pilarog.github.io/}: (1) synthetic utterances heard in isolation; and (2) synthetic utterances heard after a natural rendition of the correct preceding utterance. The second stage only evaluated the proposed system that obtained the highest naturalness score in the first stage.

Both listening tests were implemented on the Qualtrics \footnote{https://www.qualtrics.com/} platform and native speakers of English were recruited and paid through the crowd-sourcing service Prolific Academic\footnote{https://www.prolific.co/}.

\subsection{Naturalness of utterances heard in isolation}

\subsubsection{Listening test}

Hypothesis 1: \textit{Using acoustic context during training and at synthesis time improves over the baseline, even when utterances are heard in isolation (i.e., without the listener hearing the acoustic context).} 

The 5 systems tested were: (1) \textbf{Reference} ( natural speech vocoded using Waveglow) ; (2) \textbf{Baseline}; (3) \textbf{ACE only}; (4) ACE with \textbf{Order Task}; (5) ACE with \textbf{Next Task}.

We included \textbf{ACE only} to measure the benefit of also minimising the loss on the additional task (\textbf{Order Task} or \textbf{Next Task}), compared to simply providing the embedding of acoustic context as an additional input and minimising only the TTS loss. 20 utterances were randomly selected from the evaluation subset. For all systems except \textbf{Baseline}, the input to the Acoustic Context Encoder is the ground truth speech for utterance $N-1$.

We employed a MUSHRA-like listening test design in which all 5 versions of an isolated utterance were presented side-by-side. 20 participants were asked to score the naturalness of each version on a scale from 0 to 100; they also had to find the hidden \textbf{Reference} (vocoded natural speech) and give only that version a score of 100.

\subsubsection{Results}

\begin{figure}
    \centering
    \includegraphics[width=0.5\textwidth]{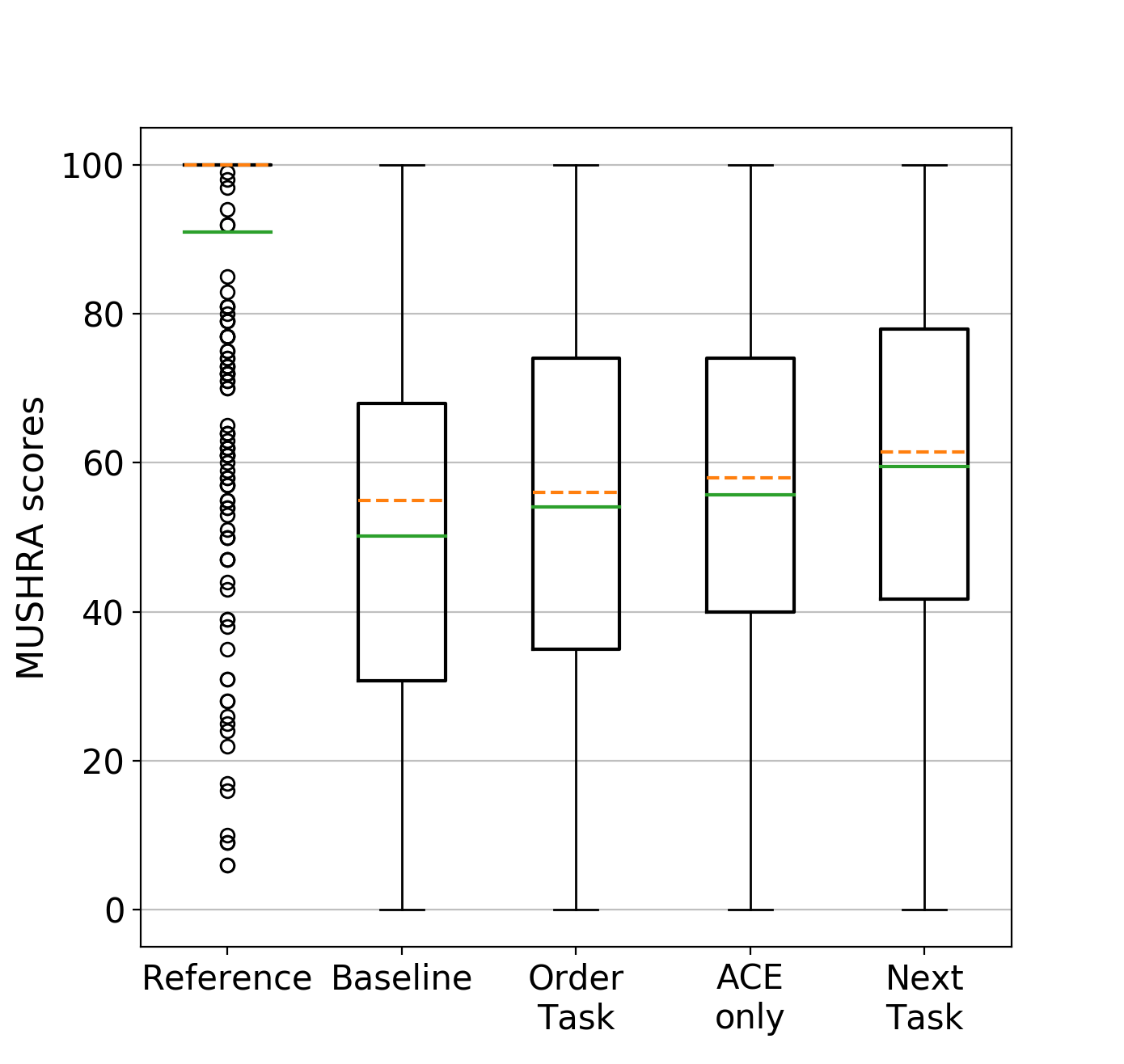}
    \caption{MUSHRA-like listening test results for the naturalness of utterances heard in isolation. In the boxes extends given the interquartile range of the data. The solid line is the mean; the dashed line is the median. Circles represent outliers.}
    \label{fig:results1}
\end{figure}

Listeners' scores were not normally distributed (Shapiro-Wilk test) and therefore we compared the means of each system using a Wilcoxon signed-rank test, and tested for statistical significance after Bonferroni correction (alpha=0.05). All the results shown in Figure \ref{fig:results1} are significantly different from one another. The \textbf{Next Task} system was scored as most natural (median: 61). The outliers (marked with circles) for \textbf{Reference} correspond to samples where participants thought that the reference was synthetic; this is plausible because some synthetic samples were very natural and high quality (including from the \textbf{Baseline}).

\subsection{Naturalness of utterances heard in context}

\subsubsection{Listening test}

The \textbf{Next Task} system was rated as the most natural in the previous section. But this could still be a ``training-only'' benefit: the acoustic context explains otherwise-unexplained variation, and so aids the model in learning the main TTS task.

To evaluate the benefit of using utterance $N-1$ \textit{at synthesis time} we conducted a pairwise forced choice preference test. The audio presented to listeners comprised the concatenation of a rendition of utterance $N-1$ and a rendition of utterance $N$ with a fixed short pause of duration 500 ms in between.

The systems compared were: 
the \textbf{Next Task} system using a natural rendition of the correct utterance $N-1$ as context;
the Next Task system using a natural rendition of a random utterance (i.e., not utterance $N-1$) as context, denoted \textbf{Next Task (random context)};
the \textbf{ACE only} system using a natural rendition of the correct utterance $N-1$ as context;
a vocoded natural \textbf{Reference}. 

Hypothesis 2: \textit{Preferences will be as follows: (1) \textbf{Next Task} is preferred over \textbf{ACE only}; (2) \textbf{Next Task} is preferred over \textbf{Next Task (random context)}; (3) \textbf{Reference} is preferred over \textbf{Next Task} although not necessarily by a large amount, given the results in the previous test.}

We manually selected 10 pairs of sentences from the test data for this task using the following criteria: the combined duration of the two sentences was not longer than 10 seconds; the combined text was meaningful. We recruited 30 participants, who were given exactly the instructions below:

\begin{itemize}
    \item You will be presented with two audios at a time: both of the same sentence.
    \item The first part of both audios is identical, however, you will notice that the second part will be slightly different.
    \item We want you select, for each pair, which second part is the most natural sequel to the first.
\end{itemize}

The text for the two sentences was presented, with the second sentence in bold face. Participants were forced to select one of the two options; there was no neutral or ``no preference'' option.

\subsubsection{Results}

\begin{figure}
    \centering
    \includegraphics[width=0.4\textwidth]{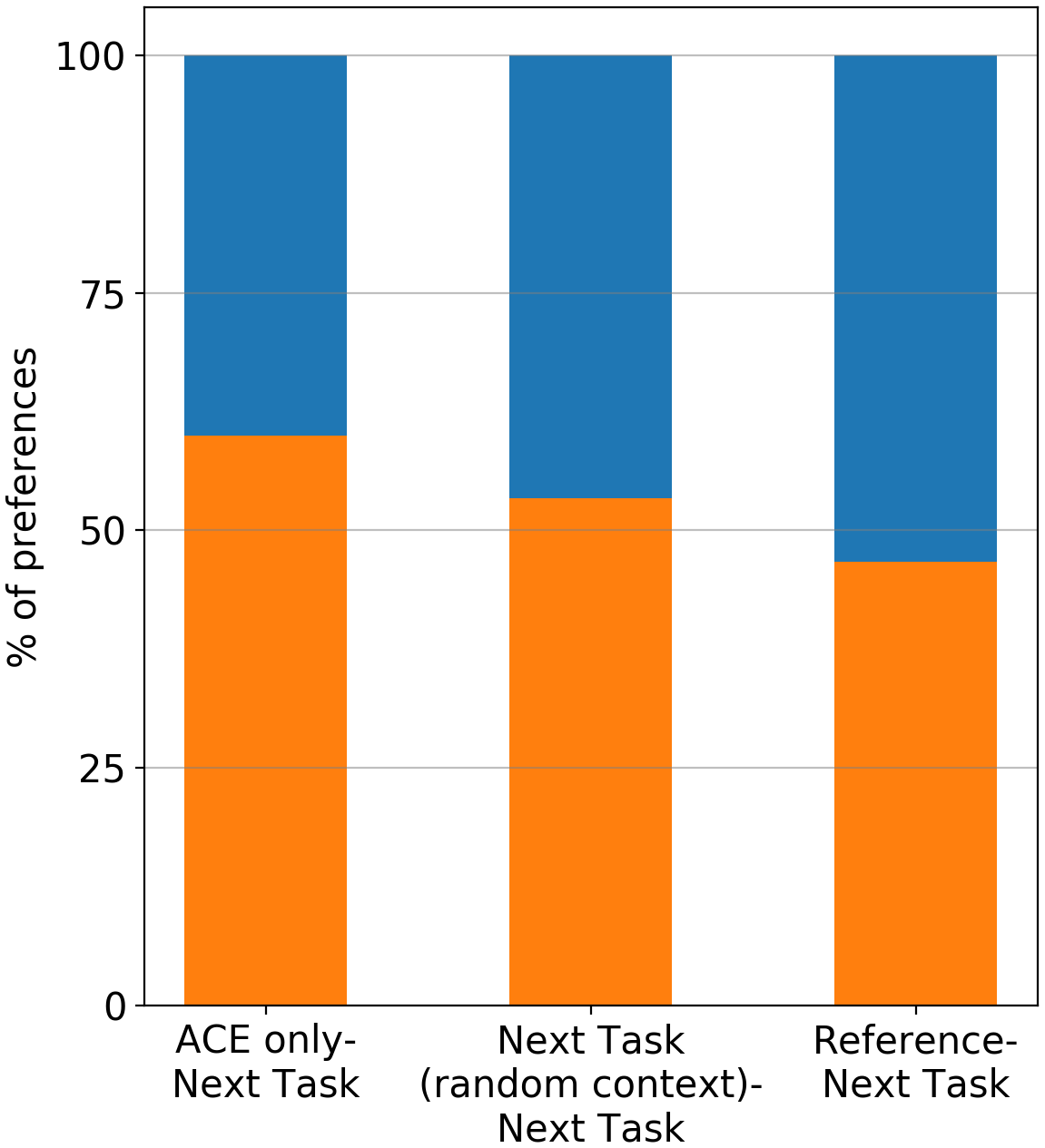}
    \caption{Pairwise forced choice results. \textbf{Next Task} is always the bottom bar.}
    \label{fig:results2}
\end{figure}

As Figure \ref{fig:results2} shows, the results follow a trend that is consistent with Hypothesis 2. We tested for significance using a binomial test, where only the \textbf{ACE only} and \textbf{Next Task} preference results were statistically significant (p-value=0.0006). In the next section we expand on the difficulties of measuring the benefits for utterances heard in context.

\section{Analysis and Discussion}

We speculate that \textbf{Order Task} didn't yield as good results as \textbf{Next Task} because, if two consecutive utterances are extremely similar in embedding space, it will be difficult to distinguish their order. The classifier converged quite quickly to an accuracy of about 65\%, which is not much better than chance (50\%). A better approach might have been to use pairs of utterances at a variety of distances apart within the data, and have the model predict that distance. 

\subsection{Qualitative evidence that the \textbf{Next Task} model uses the acoustic context to vary its output}

Regarding our best model, \textbf{Next Task}, we conducted a post-evaluation experiment to gather more evidence about the model's use of acoustic context at synthesis time.

We trained an additional model with the same architecture and loss functions as \textbf{Next Task}, but in which the extra regression task was to predict the embedding of utterance $N$ given the embedding of a random other utterance (not utterance $N-1$). This, of course, should not be possible: the model should learn to ignore the provided acoustic context.
 
\begin{figure}
    \centering
    \includegraphics[width=0.5\textwidth]{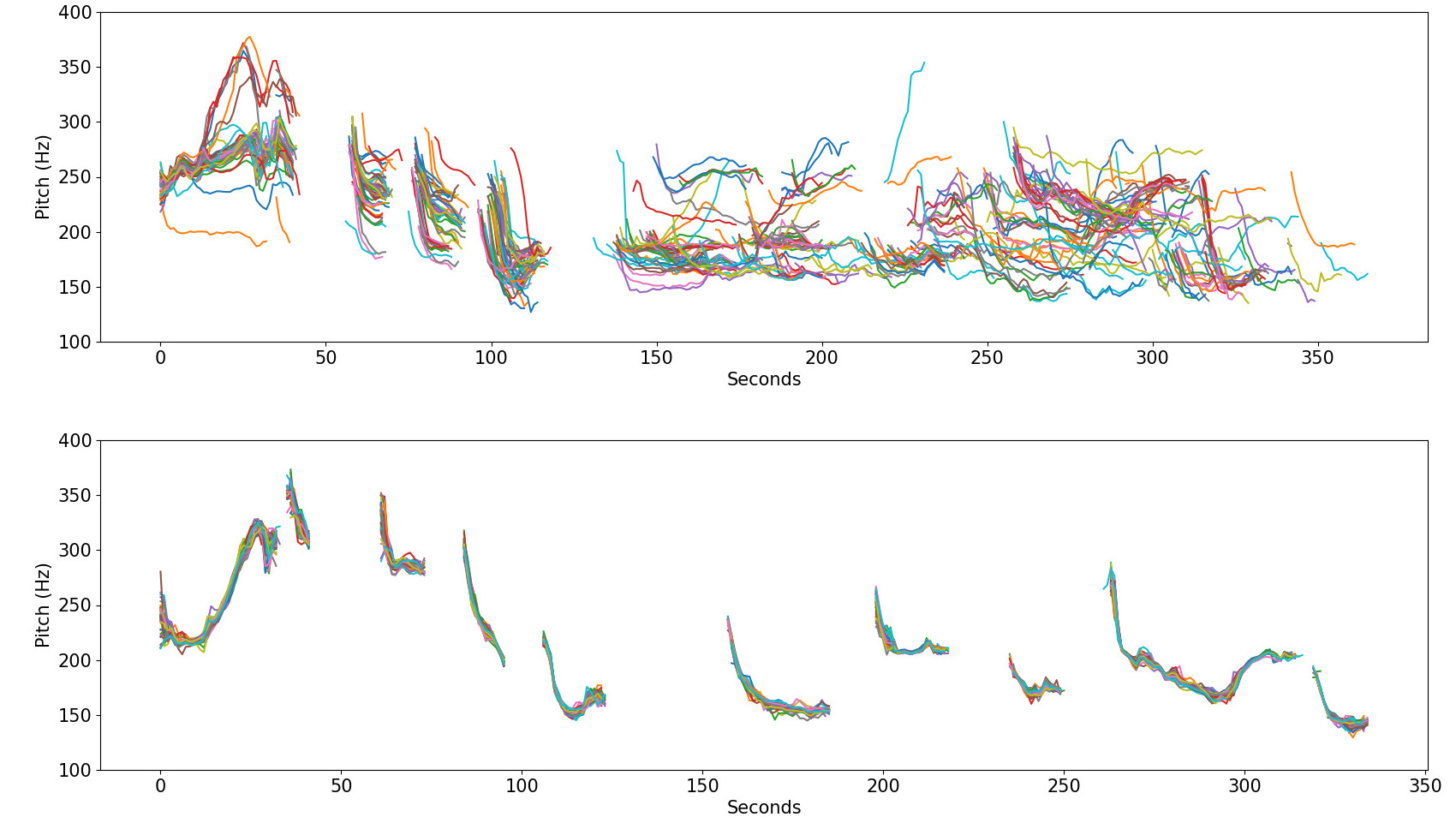}
    \caption{Pitch contours for a single text synthesized with 50 different inputs to the Acoustic Context Encoder. Upper pane: output for the \textbf{Next Task} model trained with the correct acoustic context. Lower pane: output for the same model architecture when trained with random (i.e., irrelevant) acoustic context.}
    \label{fig:analysis}
\end{figure}

We then synthesized one utterance 50 different times, using different inputs for the Acoustic Context Encoder, for each model in turn. Figure \ref{fig:analysis} shows the pitch contours for the different versions of the same utterance for each model. The model trained with random context generates only one version, regardless of the changing context: it has learned to ignore the irrelevant acoustic context provided during training and thus ignores it at synthesis time. \textbf{Next Task} generates many versions: it has learned to use the provided acoustic context during training and varies its output accordingly at synthesis time. We only plot pitch contours here, but note that the model also varies duration and pausing.

\subsection{Evaluating utterances heard in context}

This proved to be difficult. We tested the simplest case by providing listeners with only one utterance as context. This was not just to be consistent with the models being evaluated, but also because of concerns over the cognitive load imposed on participants asked to listen to and rate longer audio samples. Additionally, even given some acoustic context, there is still likely to be more than one acceptable rendition. Therefore, the forced choice task may not have been the best approach.

\section{Conclusions and Future Work}

In this paper we presented a method to leverage the sequential nature of speech data such as audiobooks or other monologues, by providing acoustic context to the model. This significantly improved naturalness of utterances heard in isolation and may also improve their contextual naturalness when heard in context.

Our approach employs an existing method for embedding speech signals (Global Style Tokens) but uses the embedding in a novel way via an extra task. Training is performed to minimise the sum of the usual TTS loss plus the loss for this extra task. A regression task (\textbf{Next Task}) proved the most successful in our experiments so far.

In this work we only use ground truth context (i.e., natural renditions) as input to the ACE. A natural next step would be to synthesise a complete sequence of utterances, using the previous synthetic audio as acoustic context when generating the current utterance.

Evaluation may be the main challenge, because we would like to present listeners with longer passages of entirely synthetic speech, such as whole paragraphs. We would also examine whether the prosody contains patterns observed in real data \cite{farrus2016paragraph}. 

We plan to explore other architectures for the Acoustic Context Encoder, such as the use of VAEs as in \cite{tyagi2019dynamic}. Any architecture that can capture global characteristics of an utterance should be applicable. We want to test how much effect the architecture of the ACE has, noting that the GST module may be over-complex for this task. 

It was the use of the extra regression task and its loss function that improved the model, not just the provision of utterance embeddings as input. This motivates the inclusion of further tasks that capture the sequential nature of the data, and experimentation with architectures for the extra task models.

Our proposed method would be suitable for any dataset that was elicited in a sequential fashion. We hope to test it with other sources of data where prosodic effects are more diverse than in LJ speech (which comprises readings of rather uninteresting non-fiction), as well ad being more consistent and meaningful. This suggests the use of  monologues performed by highly-proficient and engaging speakers. Beyond monologue, we hope to extend to conversational or other dialogue speech, and attempt to model the contextual dependency between speaker turns. This idea has been suggested \cite[for example]{hirschberg2002communication} and could now be explored using our method.

\bibliographystyle{IEEEbib}
\bibliography{text}

\end{document}